\pdfoutput=1
\documentclass[11pt]{article}
\usepackage{coling}
\usepackage{svg}
\usepackage{times}
\usepackage{booktabs}
\usepackage{siunitx}
\usepackage{caption}
\usepackage{tabularx}
\usepackage{multirow}
\usepackage{latexsym}
\usepackage{algorithm}
\usepackage{algpseudocode}
\usepackage[T1]{fontenc}
\usepackage{microtype}

\usepackage{inconsolata}

\usepackage{graphicx}
\usepackage{changepage}
\usepackage{amsmath}

\usepackage{multirow}
\usepackage{wrapfig, lipsum} 
\usepackage{float}

\usepackage[utf8]{inputenc} % allow utf-8 input
\usepackage[T1]{fontenc}    % use 8-bit T1 fonts
\usepackage{hyperref}       % hyperlinks
\usepackage{url}            % simple URL typesetting
\usepackage{booktabs}       % professional-quality tables
\usepackage{amsfonts}       % blackboard math symbols
\usepackage{nicefrac}       % compact symbols for 1/2, etc.
\usepackage{microtype}      % microtypography
\usepackage{xcolor}         % colors
\usepackage{enumitem}
\usepackage{wrapfig}
\usepackage{authblk}
\usepackage{siunitx}
\title{EffiQA: Efficient Question-Answering with Strategic Multi-Model Collaboration on Knowledge Graphs}

\author{
 \textbf{Zixuan Dong},
 \textbf{Baoyun Peng},
 \textbf{Yufei Wang},
 \textbf{Jia Fu},
\\
 \textbf{Xiaodong Wang},
 \textbf{Yongxue Shan},
 \textbf{Xin Zhou},
\\
  \textbf{National University of Defense Technology}
\\
 \small{
   \textbf{Correspondence:} \href{mailto:dongzixuan18@nudt.edu.cn}{dongzixuan18@nudt.edu.cn}
 }
}

\begin{document}
\maketitle
\begin{abstract}
While large language models (LLMs) have shown remarkable capabilities in natural language processing, they struggle with complex, multi-step reasoning tasks involving knowledge graphs (KGs). Existing approaches that integrate LLMs and KGs either underutilize the reasoning abilities of LLMs or suffer from prohibitive computational costs due to tight coupling. 
To address these limitations, we propose a novel collaborative framework named EffiQA that can strike a balance between performance and efficiency via an iterative paradigm. EffiQA consists of three stages: global planning, efficient KG exploration, and self-reflection. Specifically, EffiQA leverages the commonsense capability of LLMs to explore potential reasoning pathways through global planning. Then, it offloads semantic pruning to a small plug-in model for efficient KG exploration. Finally, the exploration results are fed to LLMs for self-reflection to further improve global planning and efficient KG exploration.
Empirical evidence on multiple KBQA benchmarks shows EffiQA's effectiveness, achieving an optimal balance between reasoning accuracy and computational costs. We hope the proposed new framework will pave the way for efficient, knowledge-intensive querying by redefining the integration of LLMs and KGs, fostering future research on knowledge-based question answering. 
\end{abstract}

\section{Introduction}

Large language models (LLMs) \citep{radford2019language,ouyang2022training,touvron2023llama1} have shown impressive capabilities across various natural language processing tasks, generating coherent and context-sensitive responses that demonstrate deep linguistic insights \citep{wei2022chain}. However, they struggle with complex, multi-step reasoning tasks, including complex arithmetic \citep{wang2022self}, commonsense \citep{zhao2024large}, symbolic reasoning \citep{pan2023logic}, and multi-hop question answering \citep{yasunaga2021qa,pan2024unifying}.
\begin{figure}[t]
  \centering
  \includegraphics[width=0.8\linewidth]{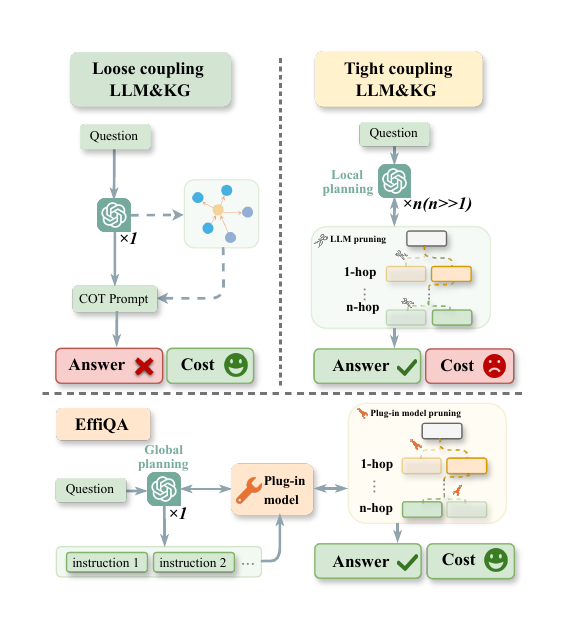} 
  \caption{LLM-based KBQA includes three paradigms: loose-coupling (LLM-only with prompts), tight-coupling (LLM exploring KG iteratively), and moderate-coupling (LLM for planning, plug-in model for KG exploration).}
  \label{fig:motivation}
\end{figure}

To enhance the reasoning abilities of LLMs, techniques like chain-of-thought (CoT) prompting \citep{wei2022chain} have been developed, enabling step-by-step rationale generation before arriving at the final answer. Despite improved performance on various reasoning tasks \citep{yao2024tree,besta2024graph,turpin2024language}, CoT prompting sometimes fails to generate sufficient sub-questions to gather all necessary information, leading to issues like hallucinations \citep{lyu2023faithful,lin2021truthfulqa}, opaque reasoning \citep{suzgun2022challenging}, and reliance on outdated data \citep{borgeaud2022improving,izacard2023atlas}.

To address these challenges, researchers have explored integrating external knowledge sources, such as knowledge graphs (KGs) \citep{sun2021jointlk,yasunaga2021qa,zhang2022greaselm,li2023chain,pan2024unifying}, into the reasoning process. These methods typically involve retrieving information from KGs, augmenting the prompt, and feeding it into LLMs. However, these loose-coupling paradigms often rely on simple data retrieval, failing to harness the full reasoning potential of LLMs. Consequently, their success hinges on the completeness and quality of the KGs.

Recent approaches have explored tighter integration between LLMs and KGs, such as Think-on-Graph (ToG) \citep{sun2023think}, Reasoning-on-Graph \citep{luo2023reasoning}, and Chain-of-Knowledge \citep{li2023chain}. These methods use LLMs to iteratively explore entities and relations in KGs, achieving better performance but at the cost of excessive computational resources and the risk of inefficiency due to potential path explosion, which can lead to suboptimal reasoning in dynamic queries.

Striking a balance between the coupling degree of LLMs and KGs, such that it can fully utilize their respective capabilities to enhance KBQA performance while maintaining efficiency, remains a challenge. To address this challenge, we propose a novel collaborative framework EffiQA for LLM-based KBQA, which leverages the commonsense capability of LLMs for global planning while offloading semantic pruning to a small plug-in model. 

Specifically, EffiQA consists of the following stages: 

\paragraph{Global Planning.} In this stage, LLM is employed to decompose the question into several semantically coherent trajectories and generate exploration instructions for exploring potential reasoning pathways and extending search space beyond the knowledge graph’s structural limits.
\paragraph{Efficient KG exploration.} In this stage, a plug-in model is employed for semantic pruning based on global planning to remove irrelevant nodes and paths during the KG search process.
\paragraph{Self-reflection.} In this stage, LLM will proceed to self-reflect on the exploration results to refine the global planning, leading to improved planning and exploration in the subsequent iteration.

To strike a balance between performance and efficiency, EffiQA introduces an enhanced querying strategy that tightly couples the LLM's instructions with constrained semantic pruning of the KG. This allows EffiQA to selectively expand the most promising graph regions based on semantics and type of entities, substantially reducing the search space while maintaining relevance. A fine-grained semantic matching process further focuses the pruning on conceptually relevant relationships.
By having the LLM provide high-level guidance while offloading computationally expensive KG traversal to a specialized model, EffiQA achieves a balanced integration. 
Different from previous methods, this collaborative approach can not only enhance the reasoning performance by combining the strengths of LLMs and KGs but also improve operational efficiency. Figure \ref{fig:motivation} shows the difference between the proposed method and previous methods. By striking a balanced integration between LLMs and KGs, EffiQA redefines the standards for efficient, knowledge-intensive querying in KBQA tasks.

\begin{figure*}[t]
  \centering
  \includegraphics[width=\textwidth]{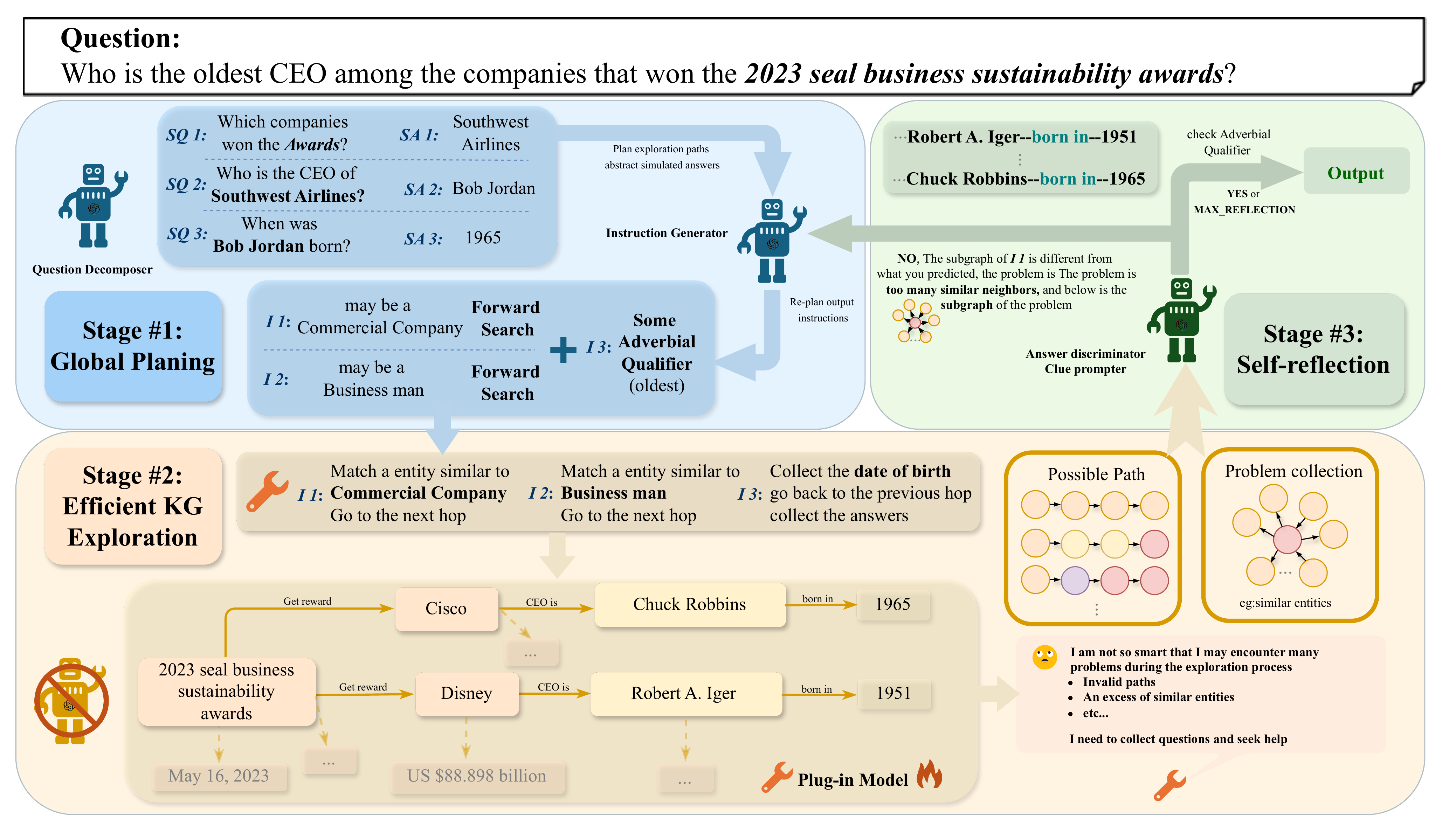} 
  \caption{The EffiQA workflow consists of three stages. First, the LLM decomposes the problem and generates instructions that include simulated answers and actions based on the problem’s logic. Next, EffiQA employs a plug-in model to execute these instructions, perform efficient knowledge graph exploration, and identify potential issues. Finally, the LLM reviews the identified problems, iteratively replans, and produces answers once sufficient information is available.}
  \label{fig:pipeline}
\end{figure*}

\section{Related Work}
\subsection{Integration of External Knowledge Sources}
Recent advancements in LLMs have focused on enhancing reasoning capabilities by integrating external information \citep{guo2023knowledgenavigator, 10417790}. Notable examples include  BlenderBot3\citep{shuster2022blenderbot} and Atlas\citep{izacard2023atlas}, which achieves marked enhancements in its performance on the Knowledge Intensive Language Tasks (KILT) benchmark \citep{petroni2020kilt}. These developments illustrate a growing trend towards the dynamic incorporation of external data sources into LLMs, which serves to enrich their foundational knowledge base and diminish the frequency of inaccuracies in generated content\citep{baek2023knowledge, 10387715}.

\subsection{Knowledge-Enhanced Reasoning in LLMs}
LLMs enhanced with structured external knowledge have shown immense potential for accurately understanding user intentions \citep{jiang2023structgpt}. Nevertheless, these models often struggle with complex reasoning tasks, such as multi-hop Knowledge Base Question Answering (KBQA), primarily due to their limited capability in decomposing multi-step problems into essential intermediate steps needed to derive an answer \citep{guan2024mitigating}. In response to this challenge, the Chain of Thought (CoT) method was introduced \citep{wei2022chain}, further developed into variations such as Auto-CoT and Zero-Shot-CoT \citep{zhang2022automatic,kojima2022large}. This method exemplifies a structured prompting technique that significantly enhances the efficiency of LLMs in navigating complex reasoning tasks. Concurrently, innovative frameworks like Chain-of-Knowledge (CoK) \citep{li2023chain} and Think-on-Graph (ToG) \citep{sun2023think} have been developed, integrating knowledge retrieval directly into the reasoning process, thereby enriching the depth and improving the factual accuracy of the generated responses.

\subsection{Advanced Frameworks for Structured Knowledge Utilization}
% \vspace{-0.8em}
The evolution of knowledge integration strategies has significantly advanced the development of frameworks that facilitate more profound interactions between LLMs and knowledge graphs. For instance, a work\citep{pan2023large} has introduced a method in which LLMs employ a greedy search algorithm to navigate KGs, enabling more nuanced interactions with data. In a related development, the Clue-Guided Path Exploration (CGPE)\citep{tao2024clue} framework, which effectively combines a knowledge base with an LLM, using clues from queries to guide systematic exploration, which in turn helps reduce computational loads. Concurrently, the Verification and Editing (VE) framework has been developed by \citep{zhao2023verify}. This framework seeks to refine the reasoning outputs of LLMs, which consequently increases both the fidelity and reliability of the model's responses.
% The evolution of knowledge integration strategies has significantly advanced the development of frameworks that facilitate more profound interactions between LLMs and knowledge graphs \citep{pan2023large}. For instance, [Jiang et al.2023] have introduced a method in which LLMs employ a greedy search algorithm to navigate KGs, enabling more nuanced interactions with data. This method notably improves the dynamic utilization of structured knowledge, thereby enhancing the accuracy and explainability of the reasoning processes. In a related development, the Clue-Guided Path Exploration (CGPE)\citep{tao2024clue} framework, which effectively combines a knowledge base with an LLM, using clues from queries to guide systematic exploration, which in turn helps reduce computational loads. Concurrently, the Verification and Editing (VE) framework has been developed by \citep{zhao2023verify}. This framework seeks to refine the reasoning outputs of LLMs. It integrates retrieval systems that verify and amend generated content, which consequently increases both the fidelity and reliability of the model's responses.

\section{Method}
The EffiQA framework consists of three main components that are executed iteratively: global planning, efficient KG exploration, and self-reflection. 
In the global planning stage, LLM is leveraged to decompose the input question into semantically coherent trajectories and generate exploration instructions, which helps explore potential reasoning pathways and extend the search space beyond the structural limits of the KG. Then, a small plug-in model is employed for efficient KG exploration, performing breadth-first search and semantic pruning on KG with the help of exploration instructions. In the self-reflection stage, LLM will reflect on the exploration results to revamp global planning and KG exploration for further improvement. Figure \ref{fig:pipeline} shows the overall framework of the proposed method.
 
% EffiQA adopts a hybrid reasoning strategy that combines a large language model (LLM) with a pre-trained language model (PLM) to enhance the system's reasoning capabilities. Specifically, it prompts LLM to perform global path exploration planning, and at the same time generates exploration instructions on the graph, executes instructions with the help of PLM capabilities, performs breadth-first search and semantic pruning on KG, and finally uses LLM to complete inference path analysis and aggregation. As shown in Figure 2, EffiQA mainly consists of 3 stages: global planning, search execution, answer aggregation and the iterative interaction process between them.
\subsection{Global Planning}
% \vspace{-0.4em}
In the global planning phase, the LLM leverages its powerful reasoning abilities to globally plan the exploration pathway. At this stage, LLM will give a set of instructions based on the given question and initial entity to guide the plug-in model to explore the graph.

\paragraph{Query Decomposition}  Initially, the LLM identifies the primary subject entity \(e_0\) from the query and deconstructed into \(M\) sub-questions \(Q = \{q_1, q_2, \ldots, q_M\}\) with adverbial qualifiers \(R = \{r_1, r_2, \ldots, r_N\}\), where ($N\leq M$). This process is formulated as:\(Q, R \leftarrow \text{Decompose}(query, e_0)\).

\paragraph{Instruction Generation and Optimization}  After decomposition, the LLM generates simulated answers \(A_i \leftarrow \text{Simulate}(e_i, q_i, r_i)\) and constructs corresponding instructions \( \text{instructions} \leftarrow \text{S}(e_i, q_i, r_i) \) to guide the plug-in model in efficiently navigating the knowledge graph (KG). Specifically, "forward search" instructions are created for sub-questions, and "adverbial qualifier" instructions are designed to handle qualifiers; these are collectively referred to as action instructions. Additionally, we have developed other action instructions tailored for the plug-in model and generate the appropriate matching instructions. For example, Figure \ref{fig:case} illustrates the retrieval of a sports event type.

\subsection{Efficient KG exploration}
The simulated answers and actions generated by Global Planning can effectively guide the plug-in model based on an intuitive prior premise: when LLM answers a question, the content of the answer may be inaccurate, but will generate simulated answers of the same type or with similar semantics as the answer. Using the simulated answers, we implement a constrained search with semantic pruning, selectively expanding promising entities to optimize efficiency.
\begin{algorithm}[ht]
\caption{EffiQA Framework}
\begin{algorithmic}[1]
\State \textbf{Input:} Query $q$, Knowledge Graph $G$
\State $e_0 \leftarrow$ \textsc{ExtractEntity}($q$)
\While{not converged}
    \State $Q, R \leftarrow$ \textsc{DecomposeQuery}($q$, $e_0$)
    \State $A \leftarrow$ \textsc{SimulateAnswers}($Q$, $R$)
    \State $\text{Instr} \leftarrow$ \textsc{GenerateInstr}($e_0$, $Q$, $R$)
    \For{each instr in $\text{Instr}$}
        \State $Paths, Problems\leftarrow$\textsc{KGExplore}($G$, $e_0$, instr, $A$)
    \EndFor
    \State $Result,Clues\leftarrow$\textsc{Aggregate}(Paths, Problems)
    \If{Result satisfactory}
        \State \textbf{Output:} Final Answer
        \State \textbf{break}
    \Else
        \State \textsc{RevisePlan}($Clues$)
    \EndIf
\EndWhile
\end{algorithmic}
\end{algorithm}

\paragraph{Initialization and Systematic Exploration}  Graph initialization define the knowledge graph \( G = (E, R) \) and start BFS from \(e_0\) and for each entity \(e_{\text{current}}\) in BFS, evaluate relations \(r\), select a representative tail entity, and perform semantic matching with simulated answers using entity descriptions and triples.

\begin{figure}[ht]
 \centering
 \includegraphics[width=\linewidth]{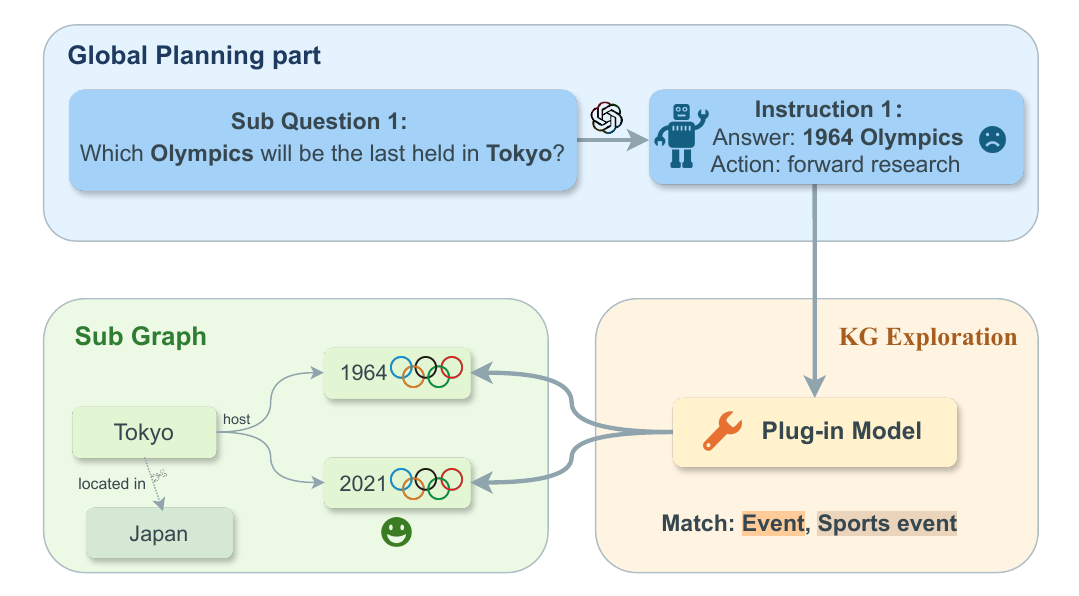}
 \caption{Despite the challenges associated with  the temporal constraints of the training corpus (up to March 2021), the plug-in model effectively executes accurate match pruning. The detailed entity typing depicted in the figure serves solely for illustrative purposes}
 \label{fig:case}
\end{figure}

\paragraph{Semantic Matching and Graph Traversal}   EffiQA uses fine-grained entity typing (FGET) based on the Ontonotes dataset \citep{dan2014context} for semantic matching, ensuring exploration focuses on pertinent relations despite potential inaccuracies
For example, as shown in Figure \ref{fig:case}, even if inaccurate instruction content is given due to large model time constraints, plug-in model execution instructions still allow the search for premises that are conceptually consistent despite being inaccurate in the simulation Continuing below, we demonstrate the robustness of the semantic matching process in managing time-sensitive inaccuracies by focusing on categorical relevance rather than precise data accuracy.

% \begin{figure}[t]
%   \centering
%   \includegraphics[width=1\linewidth]{case.png} 
%   \caption{Neither the illusion of large models nor the issue of corpus time limit will prevent PLM from making correct matching pruning(It is assumed that the training LLM corpus is as of March 2021).
%   \label{tab:case}
%   }
% \end{figure}

If a representative entity \(A'_{\text{rep}}\) matches \(A_i\), extend exploration to all entities linked by \(r_i\), controlling branching via a threshold. Unmatched relations \(r\) are pruned. Each entity is also checked against adverbial qualifiers \(R_i\). 
Upon reaching the reasoning depth, the search ends and qualifying paths are collected. Problems encountered (e.g., unmatched paths, excessive branching) are collected with their subgraphs for further processing.

% \vspace{-0.3em}
\subsection{Self-reflection}

\paragraph{Detailed Answer Aggregation}
In this phase, the LLM synthesizes results from efficient KG exploration to ensure answers are contextually relevant and precisely address the query. Each reasoning path is verified against its adverbial qualifiers \(R_i\), and terminal entities are checked against the final constraints \(Q_{\text{final}}\). When multiple initial entities \(E_0\) are present, validated paths are intersected to identify consistently relevant entities, enhancing answer robustness.

\paragraph{Strategic Replanning and Iteration}
Problematic paths are flagged for re-evaluation and replanning \(R_{\text{plan}}\). During this process, the LLM reflects on issues identified in the efficient KG exploration’s subgraphs and the generated clue prompts. It then performs a global replanning that integrates the problematic subgraphs and clue prompts, ensuring comprehensive coverage and addressing reasoning gaps. This iterative approach systematically resolves inaccuracies and inefficiencies, leading to more robust final answers. The final selection is based on intersected entities \(E'_{\text{final}}\), ensuring the answer is derived from validated paths and fully meets the initial query requirements.

%%%～～～～～～～～～～～～～～～～～～～～～～～～～～～～～～～～～～～～～～～～～～～
%%%～～～～～～～～～～～～～～～～～～～～～～～～～～～～～～～～～～～～～～～～～～～
% \vspace{-0.4em}
\section{Experiments}\label{sec:exp}
% \vspace{-0.4em}
\subsection{Experimental Setup}
\subsubsection{Datasets and Metrics}
To rigorously evaluate the performance of the EffiQA framework, we selected five benchmark datasets: Complex Web Questions (CWQ) \cite{talmor2018web}, WebQuestionSP (WebQSP) \cite{yih2016value}, GrailQA \cite{gu2021beyond}, QALD10-en \cite{perevalov2022qald}, and Simple Questions \cite{bordes2015large}. These datasets were chosen for their varying complexity and the different types of reasoning challenges they present, ranging from simple fact retrieval to sophisticated multi-hop questions. We assessed the framework on two primary metrics: cost efficiency, which measures the resource consumption per query, and answer accuracy (Hits@1), defined by the precision of responses compared to the gold standards. At the same time, since the recall rate is crucial during path exploration, incorrect pruning will cause errors in path exploration and lead to wrong answers. Therefore, to retain the correct path when exploring the knowledge graph without incorrect pruning, we define pruning recall as one of the important indicators of its path exploration. When calculating efficiency, we use the average question consumption of floating-point operations (FLOPs) as an indicator of computational overhead. Cost efficiency is defined as the ratio of the pruning recall rate to FLOPs.

% To rigorously evaluate the performance of the EffiQA framework, we selected five benchmark datasets: Complex Web Questions (CWQ)\citep{talmor2018web}, WebQuestionSP (WebQSP)\citep{yih2016value}, GrailQA\citep{gu2021beyond}, QALD10-en\citep{perevalov2022qald} and Simple Question\citep{bordes2015large}. These datasets were chosen for their varying complexity and the different types of reasoning challenges they present, from simple fact retrieval to sophisticated multi-hop questions. We assessed the framework on two primary metrics: cost efficiency, which measures the resource consumption per query, and answer accuracy(Hits@1), defined by the precision of responses compared to the gold standards. At the same time, since the recall rate is very important during path exploration, wrong pruning will cause errors in path exploration and lead to wrong answers. Therefore, in order to retain the correct path when exploring the KG without pruning it incorrectly, we define pruning recall as one of the important indicators of its path exploration. When calculating efficiency, we use the average question consumption of floating point operations (FLOPs) as an indicator of computational overhead. Cost efficiency is defined as the ratio of pruning recall rate to FLOPs.

%%%实验1，整体效果，包含几个数据集，chatgpt，gpt4，deepseek三个模型基座
\begin{table*}[ht]
  
  \centering
  \resizebox{\textwidth}{!}{ % Adjust table width to fit the page
  \begin{tabular}{lccccc}
    \toprule
    \multirow{1}{*}{Method} & \multirow{1}{*}{CWQ} & \multirow{1}{*}{WebQSP} & \multirow{1}{*}{GrailQA} & \multirow{1}{*}{QALD10-en} & \multirow{1}{*}{Simple Questions}   \\
    \midrule
    \multicolumn{6}{c}{Without external knowledge} \\
    \midrule
    IO prompt w/GPT-3.5-turbo\textsuperscript{$\alpha$} & 37.6 & 63.3 & 29.4 & 42.0 & 20.0   \\
    CoT w/GPT-3.5-turbo\textsuperscript{$\alpha$} & 38.8 & 62.2 & 28.1 & 42.9 & 20.3   \\
    SC w/GPT-3.5-turbo\textsuperscript{$\alpha$} & 45.4 & 61.1 & 29.6 & 45.3 & 18.9   \\
    \midrule
    \multicolumn{6}{c}{With external knowledge} \\
    \midrule
    Prior FT SOTA & 70.4\textsuperscript{$\beta$} & 82.1\textsuperscript{$\gamma$} & 75.4\textsuperscript{$\delta$} & 45.4\textsuperscript{$\epsilon$} & \underline{85.8}\textsuperscript{$\zeta$}   \\
    Prior Prompting SOTA & - & 74.4\textsuperscript{$\eta$} & 53.2\textsuperscript{$\eta$} & - & -  \\
    Prior tight-coupling SOTA\textsuperscript{$\alpha$}  &\underline{72.5} & 82.6 & \underline{81.4} & \underline{54.7} & 66.7   \\
    % ToG w/ChatGPT & 57.1 & 76.2 & 68.7 & 50.2 & 53.6 & 54.5  \\
    % ToG w/GPT-4 & 67.6 & 82.6 & 81.4 & 53.8 & 66.7 & 57.9  \\
    % \midrule
    % \multicolumn{7}{c}{Efficient LLM QA} \\
    \midrule
    % \textbf{EffiQA (ours)} w/Deepseek-V2 & 61.7 & 67.4 & 69.1 & 50.2 & 69.4 \\
    \textbf{EffiQA (ours)} w/GPT-3.5-turbo & 52.1 & 65.2 & 63.3 & 46.2 & 65.7 \\
    \textbf{EffiQA (ours)} w/GPT-4 & \textbf{69.5} & \textbf{82.9} & \textbf{78.4} & \textbf{51.4} & \textbf{76.5} \\
    \bottomrule
  \end{tabular}
  
}
\caption{Comparison between EffiQA and related methods on KBQA tasks. Method $\alpha$~\citep{sun2023think} has a significantly higher cost compared to EffiQA. $\beta$~\citep{das2021case}, $\gamma$~\citep{yu2022decaf}, $\delta$~\citep{gu2022don}, $\epsilon$~\citep{borroto2022sparql}, $\zeta$~\citep{baek2023direct}, $\eta$~\citep{li2023few}.
}
\label{tab:comparison_methods}
\end{table*}

\subsubsection{Baselines}

EffiQA is benchmarked against key established methods, including methods of reasoning without external knowledge, such as standard prompts (IO prompts) \cite{brown2020language} and CoT prompts \cite{zhang2022automatic}, as well as some state-of-the-art methods using external knowledge reasoning. These comparisons cover hint-based methods, fine-tuning methods, and different KBQA architectures. IO prompts and CoT prompts are two methods that do not require external knowledge, whereas the tight-coupling KBQA method is the most advanced approach for KBQA. These comparisons help illustrate the necessity of external knowledge in answering complex queries and demonstrate the cost-effectiveness of EffiQA compared to existing state-of-the-art methods. Through these comparisons, the flexibility and cost-effectiveness of EffiQA across various datasets can be proven.

% EffiQA is benchmarked against key established methods, including methods of reasoning without external knowledge: standard prompts (IO prompts) \citep{brown2020language}, chain of thought prompts (CoT prompts) \citep{zhang2022automatic}, and some state-of-the-art methods using external knowledge reasoning,  These comparisons cover hint-based methods, fine-tuning methods, and different KBQA architectures. IO prompt and CoT prompt are two methods that do not require external knowledge. tight-coupling KBQA method is the most advanced method for knowledge-based question answering. These comparisons help illustrate the necessity of external knowledge in answering complex queries and demonstrate the cost-effectiveness of EffiQA compared to existing state-of-the-art methods. Through these comparisons, it can be proven that the flexibility and cost-effectiveness of EffiQA on various datasets.

\subsubsection{Experiment Details}
Comprehensive experiments on EffiQA were performed using two distinct LLMs: GPT-3.5-turbo, GPT-4\footnote{Both GPT-3.5-turbo and GPT-4 can be accessed at \url{https://openai.com/}} and four models with different parameter sizes for comparison. These models were selected to assess the framework's scalability and performance across varying computational capabilities. GPT-4 was utilized to explore the limits of performance in more complex scenarios. To ensure consistent and reproducible results, the temperature setting for all interactions with these models was fixed at 0, eliminating randomness in responses.

% Experiments on EffiQA were performed using three distinct large language models: ChatGPT, GPT-4\footnote{Both ChatGPT and GPT-4 can be accessed at https://openai.com/}, and DeepSeek-V2\citep{deepseekai2024deepseekv2}, which is a 236B MoE model with inference activation 21B parameters, each selected to assess the framework's scalability and performance across varying computational capabilities. Specifically select one latest open-source MOE model to evaluate whether it can maintain good reasoning capabilities when the number of activation parameters is small., while GPT-4 were utilized to explore the limits of performance in more complex scenarios. To ensure consistent and reproducible results, the temperature setting for all interactions with these models was fixed at 0, eliminating randomness in responses. 

The plug-in model used for node semantic matching and path pruning utilizes RoBERTa \cite{liu2019roberta}, fine-tuned on the modified OntoNotes v5 dataset\footnote{OntoNotes v5 can be downloaded at \url{https://huggingface.co/datasets/conll2012_ontonotesv5}}, and a 10,000 entity typing training set generated using GPT-4 for fine-tuning. The dataset is generated using the classification standards from the Context-Dependent Fine-Grained Entity Type Tagging method \cite{dan2014context} as the basis for entity typing. Since only the path pruning process involves entity typing, to ensure accurate recognition of the named entity recognition (NER) part, we use special markers to mark the entities that need to be classified and then fine-tune the model to ensure that the specified entities can be correctly identified.

% The plug-in model used for node semantic matching and path pruning uses RoBERTa\citep{liu2019roberta}, finetuned on modified ontonotes v5 dataset\footnote{Ontonotesv5 can be downloaded at https://huggingface.co/datasets/conll2012\_ontonotesv5}, and 10,000 entity typing training set generated using GPT-4 for fine-tuning, The data set is generated using the classification standards in use the Context-Dependent Fine-Grained Entity Type Tagging method\citep{dan2014context} as the basis for our entity typing.. Since only the path pruning process involves entity typing, in order to ensure accurate recognition of the NER part, we use special markers to mark the entities that need to be classified and then fine-tune them to ensure that the entities specified by the classification can be correctly identified.

% \vspace{-0.4em}
\subsection{Main Results}
% \vspace{-0.4em}
\subsubsection{Comparison to Other Methods}
We compare EffiQA with some frameworks without external knowledge. As a baseline for large model capabilities, that is, the knowledge contained in the large models themselves, as can be seen from Table \ref{tab:comparison_methods}, the methods without external knowledge generally perform poorly, reflecting the LLMs dependence on knowledge graphs for answering knowledge-based questions. EffiQA leverages both large and smaller specialized language models integrated with external knowledge graphs, surpassing all methods that do not use external knowledge. At the same time, EffiQA provides a distinctive advantage over traditional fine-tuning approaches with its plug-and-play capability that requires no dataset-specific training.

% we compare with some frameworks without external knowledge. As a baseline for large model capabilities, that is, the knowledge contained in the large model itself,  As can be seen from Table \ref{tab:comparison_methods}, the method without external knowledge generally performs poorly, which reflects the dependence of LLM on the knowledge graph for knowledge question. EffiQA leverages both large and smaller specialized language models integrated with external knowledge graphs, Surpassing all methods that without external knowledge. At the same time, EffiQA provides a distinctive advantage over traditional fine-tuning approaches with its plug-and-play capability that requires no dataset-specific training. 

Comparison with methods using external knowledge shows that even without any fine-tuning, EffiQA still outperforms existing fine-tuning methods. Particularly noteworthy is its performance on single-hop datasets. In comparison to ToG, which tightly couples with LLMs and KGs, EffiQA demonstrates competitive strength. Notably, EffiQA excels in single-hop datasets like SimpleQuestions, where it achieves 65.7\% accuracy with GPT-3.5-turbo and 76.5\% with GPT-4, underscoring its effective global planning for enhancing accuracy without sacrificing recall—a common shortfall in larger models that heavily prune data. The framework's strategy for addressing simple questions enhances accuracy without compromising the recall rate, a common issue in large models that employ aggressive pruning techniques, where EffiQA's global planning proves highly effective.

% comparison with the method using external knowledge shows that even without fine-tuning with training data, EffiQA is still better than existing fine-tuning methods and simply using external knowledge prompts, Particularly noteworthy is its performance on single-hop datasets, In comparison to methods that rely solely on large language models (LLM) for all processing (tighter-coupling integration of LLMs and KGs methods), EffiQA shows competitive strength, Notably, EffiQA excels in single-hop datasets like Simple Questions, where it achieves 65.7\% accuracy with ChatGPT and 76.5\% with GPT-4, underscoring its effective global planning which enhances accuracy without sacrificing recall—a common shortfall in larger models that heavily prune data. where EffiQA's global planning proves highly effective. The framework's strategy for addressing simple questions enhances accuracy without compromising the recall rate, a common issue in large models that employ aggressive pruning techniques.

% \vspace{-1.5em}
EffiQA also demonstrates competitive results on multi-hop datasets such as ComplexWebQuestions (CWQ) and WebQuestionsSP (WebQSP), scoring 69.5\% and 82.9\% respectively when using GPT-4. These results demonstrate that EffiQA's strategy of leveraging external knowledge significantly improves LLMs deep reasoning capabilities, effectively managing complex queries that often pose challenges to tighte-coupling approaches that integrate LLMs and knowledge graphs, because they usually consume much more than EffiQA while achieving similar performance.

% EffiQA also demonstrates competitive results on multi-hop datasets such as CWQ and WebQSP, scoring 69.5\% and 82.9\% respectively when using GPT-4. These results demonstrate that EffiQA's strategy of leveraging external knowledge significantly improves LLM's deep reasoning capabilities, effectively managing complex queries that often pose challenges to tighter-coupling approaches.

\subsubsection{Performance with Different Backbone Models}
% \vspace{-1.5em}
% \noindent

In exploring the integration of EffiQA with different LLM backbones, we conducted ablation studies using a range of different scale models in table \ref{tab:performance}, Since EffiQA relies heavily on LLM with powerful reasoning capabilities to give accurate instructions, we chose Llama3.1-8B \cite{vavekanand2024llama}, GPT-3.5-turbo, Deepseek-V2, and GPT-4 as our planning and aggregation iteration modules. These studies aim to evaluate how the underlying LLM affects the overall accuracy of the system on multiple datasets such as CWQ and WebQSP.

\begin{table}[ht]
\centering
\renewcommand{\arraystretch}{1} % Adjust row height for compactness
\setlength{\tabcolsep}{4pt} % Adjust column separation for fitting table within text width
% \small % Reduce font size
\begin{tabular}{@{}l S[table-format=2.1] S[table-format=2.1]@{}}
\toprule
\textbf{Method} & \textbf{CWQ} & \textbf{WebQSP} \\ 
\midrule
\multicolumn{3}{l}{\textbf{Fine-tuned Baseline}} \\
NSM \textsuperscript{$\alpha$} & 53.9 & 74.3 \\
DeCAF \textsuperscript{$\beta$} & 70.4 & 82.1 \\
\midrule
\multicolumn{3}{l}{\textbf{Prompting Baseline}} \\
KD-CoT \textsuperscript{$\gamma$} & 50.5 & 73.7 \\
\midrule
\multicolumn{3}{l}{\textbf{LLMs}} \\
COT (Llama3.1-8B)\textsuperscript & 32.8 & 56.6 \\
EffiQA (Llama3.1-8B) & 37.4 & 58.3 \\
\textbf{Gain} & \textbf{+4.6} & \textbf{+1.7} \\
\addlinespace
COT (DeepSeek-V2) & 41.2 & 57.8 \\
EffiQA (DeepSeek-V2) & 61.7 & 67.4 \\
\textbf{Gain} & \textbf{+20.5} & \textbf{+9.6} \\
\addlinespace
COT (GPT-3.5-turbo)\textsuperscript{$\delta$}  & 38.8 & 62.2 \\
EffiQA (GPT-3.5-turbo) & 52.1 & 65.2 \\
\textbf{Gain} & \textbf{+13.3} & \textbf{+3.0} \\
\addlinespace
COT (GPT-4)\textsuperscript{$\delta$} & 46.0 & 67.3 \\
EffiQA (GPT-4) & 69.5 & 82.9 \\
\textbf{Gain} & \textbf{+23.5} & \textbf{+15.6} \\
\bottomrule
\end{tabular}
\caption{Performance comparison of methods on CWQ and WebQSP datasets. EffiQA consistently improves performance. $\alpha$~\citep{he2021improving}, $\beta$~\citep{yu2022decaf}, $\gamma$~\citep{wang2023knowledge}, $\delta$~\citep{sun2023think}.}
\label{tab:performance}
\end{table}

\begin{figure}[ht]
  \centering
  \includegraphics[width=1\linewidth]{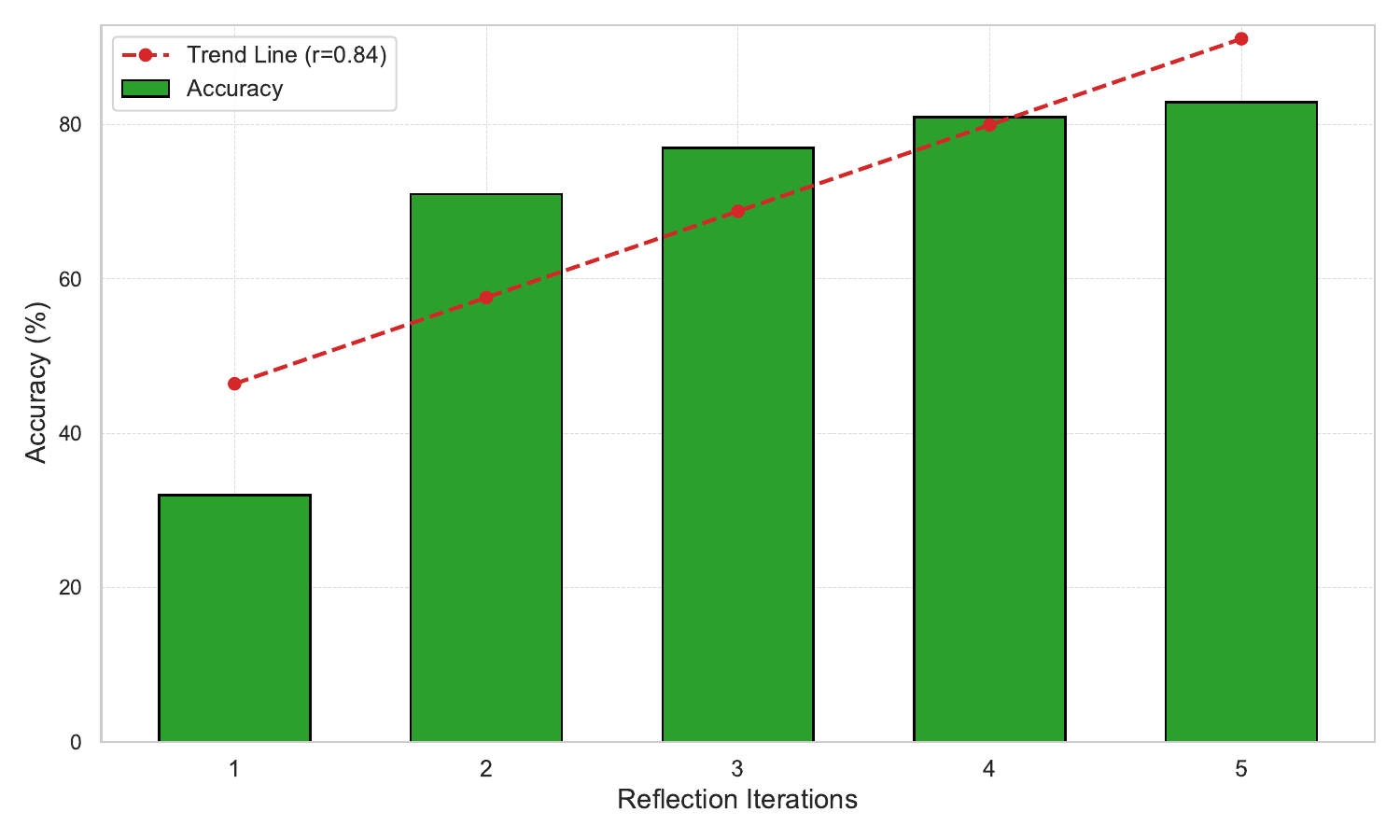} 
  \caption{Accuracy by Reflection Iterations on WebQSP Dataset (MAX\_REFLECTION=5)}
  \label{tab:reflection number}
\end{figure}

%$\alpha$~\citep{he2021improving}, $\beta$~\citep{das2021case}, $\gamma$~\citep{shu2022tiara}, $\delta$~\citep{yu2022decaf},
% The results show that the performance of EffiQA generally improves with the capacity and complexity of the adopted model. It is worth noting that although the number of parameters activated per inference of DeepSeek-V2 is not as high as that of ChatGPT, the overall performance has improved, illustrating the effectiveness of the MoE architectural model. GPT-4 has consistently outperformed other models in complex multi-hop query scenarios, leveraging its powerful inference capabilities. Notably, EffiQA is very sensitive to the performance of the LLM. When the LLM's reasoning capabilities for global planning and self-reflection increase, the overall performance of the EffiQA system improves significantly. The excellent reasoning of the LLM provides more accurate instructions and self-reflections, enhancing the plug-in model pruning accuracy in subsequent steps. This reduces the process of repeated reasoning and improves the overall accuracy rate.
\begin{figure*}[ht]
  \centering
  \includegraphics[width=1\linewidth]{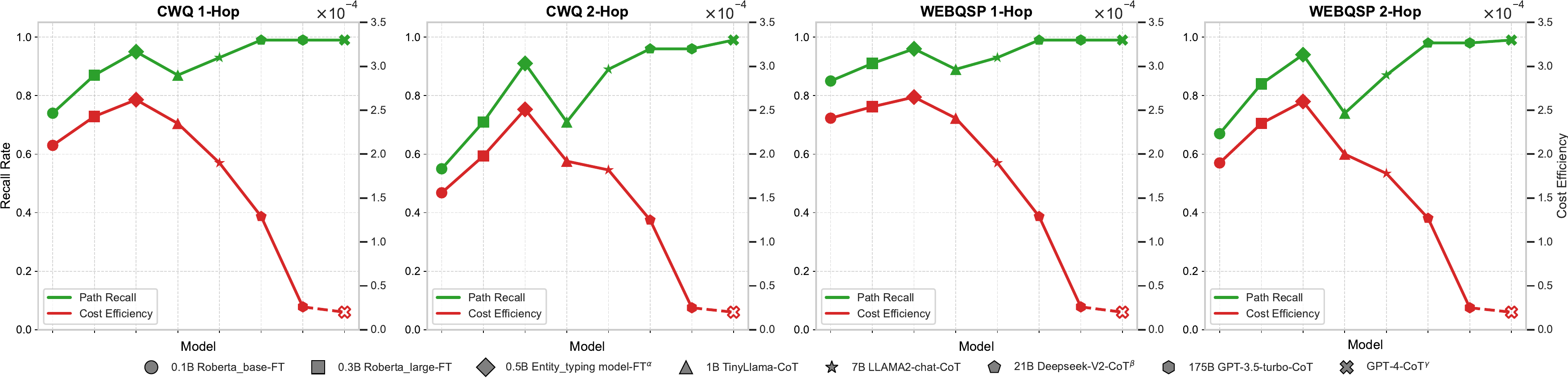} 
  \caption{Path Recall Rate and Cost Efficiency of various model sizes on CWQ and WebQSP datasets. The medium-sized entity typing model achieves an optimal balance between recall and cost. $\alpha$: RoBERTa models fine-tuned with increased parameters. $\beta$: Activation parameters calculated using the MoE model. $\gamma$: GPT-4 parameter estimates based on scale, as exact numbers are undisclosed.}
  \label{tab:Model Scale}
\end{figure*}

The results show EffiQA’s performance improves with model capacity and complexity. Notably, Deepseek-V2 activates fewer parameters than GPT-3.5-turbo yet outperforms it, demonstrating the MOE architecture’s effectiveness. GPT-4 excels in complex multi-hop queries due to strong inference capabilities. EffiQA is highly sensitive to the LLM’s reasoning abilities, such as global planning and self-reflection, which enhance accuracy by improving instructions and PLM pruning, reducing repeated reasoning. When integrated with Llama3.1-8B, EffiQA gains performance, but improvements are limited by the smaller scale, highlighting its reliance on robust planning typically found in larger models.

% \vspace{-0.4em}
\subsection{Ablation study}
% \vspace{-0.3em}
\subsubsection{Effect of Reflection Iterations on Accuracy}
To investigate the impact of reflection iterations on model accuracy, we conducted experiments on the WebQSP dataset with a maximum of 5 reflection iterations. As shown in Figure \ref{tab:reflection number}, the accuracy of the model improves consistently with an increasing number of reflection iterations. The trend line in the graph indicates a strong positive correlation (r=0.84) between the number of reflections and the accuracy rate, demonstrating that iterative reflections contribute to better performance in entity classification tasks.

The progressive improvement suggests that incorporating multiple reflection iterations enables the model to refine its understanding and make more accurate predictions, possibly by reinforcing correct classifications and learning from mistakes. This reinforces the utility of our approach, especially in scenarios where high accuracy is paramount. By employing up to 5 reflection iterations, the model achieves significant performance gains, providing a balance between computational cost and accuracy enhancement.
\subsubsection{Model scale, Efficiency and Performance}

In order to prove that when PLM obtains LLM instructions, the recall rate of semantic matching and path pruning can achieve equivalent effects to LLM direct pruning, but its cost-effectiveness has obvious advantages compared with large models, we compare Models of different sizes perform in path pruning and semantic matching tasks through entity typing. The following figure \ref{tab:Model Scale} shows the change curve of recall rate and cost-effectiveness of path pruning by models of different sizes under different hop numbers in the CWQ and WebQSP data sets.  

among which models above 1B are the current mainstream decoder-only architecture, using the generative classification method. For these models, the CoT method is used and 2-shot is used for inference, while the models below 1B are dedicated entity typing models. Inference is performed after fine-tuning. The fine-tuning process uses 2*NVIDIA 4090 24G, the learning rate is set to 5e-6, and the original data set is used plus the generated training set for fine-tuning for 20 epochs. It is worth noting that the recall rate of path pruning is not equal to the accuracy of the model's direct classification of entities, because the plug-in model always tends to classify semantically similar entities into one category. If the plug-in model is execute fine-grained entities typing and gives an incorrect result, the correct path may still be recalled successfully due to exploring the semantic similarity of entities, so the recall rate of path pruning is usually higher than the recall rate of entity typing. It can be seen from the results that some dedicated fine-tuning plug-in models have achieved performance comparable to existing large models in path matching tasks based on entity classification, even better than LLM with a small number of parameters. At the same time, their cost-effectiveness is much higher than that of large models.
% \vspace{-0.5cm}
% \subseteq
% \vspace{-0.6em}

\subsubsection{Computational Cost}
Cost analysis highlights EffiQA's ability to significantly reduce the number of queries per input question for large models, thereby reducing computational expense. This efficiency not only enhances the applicability of the framework in resource-limited environments but also emphasizes its commercial potential for scalable real-world applications. In the cost consumption experiment on the KBQA data set, we set the exploration depth and width of TOG to 3 and stopped it when the LLM requests for a problem exceeded 30 times. 

At the same time, we set the number of Self-reflections of EffiQA on single-hop problems. The threshold is set to 5, and the multi-hop value is set to 10. Tests are conducted on the CWQ and WebQSP data sets. We divide the cost consumption results of the above data sets into single and multi-hop calculations respectively. From table \ref{tab:cost}, we can see that whether it is 1-hop or multi-hop inference on KG, the number of inferences and request cost of this solution are much smaller than the existing SOTA model TOG.
\begin{table}[ht]
\centering
\begin{tabular}{@{}lcc@{}}
\toprule
Method & 1-Hop  & Multi-Hop \\
\midrule
TOG w/GPT-3.5-turbo & 16.7 & 25.6 \\
TOG w/GPT-4 & 14.8 & 21.4 \\
EffiQA w/GPT-3.5-turbo & 4.7 & 7.3 \\
EffiQA w/GPT-4 & 3.2 & 6.5 \\
\bottomrule
\end{tabular}
\caption{Average number of calls to LLM per question}
\label{tab:cost}
\end{table}
% \vspace{-0.3em}
\section{Conclusion}

In this work, we proposed EffiQA, a new integration paradigm of LLMs and KGs for multi-step reasoning. Through an iterative paradigm of global LLM planning, efficient KG exploration, and self-reflection, EffiQA balances leveraging LLM capabilities with maintaining computational efficiency. The global planning outlines promising trajectories and generates instructions to guide semantic pruning during efficient KG traversal, reducing search spaces. Exploration results then refine the global plan iteratively. Extensive experiments demonstrate EffiQA's ability to optimally balance accuracy and costs. 

\section*{Limitations}
EffiQA exhibits sensitivity to the capabilities of large models, relying on their reasoning abilities for optimal performance. This dependence means that any limitations in the large model's ability can directly affect EffiQA's outcomes. Furthermore, the plug-in model encounters performance bottlenecks when scaling to larger or more complex knowledge graphs. As the knowledge graph grows, the computational effort required for effective exploration and semantic pruning increases, potentially slowing down processing and limiting the system's efficiency in extensive datasets. These factors constrain EffiQA's scalability and adaptability in more demanding scenarios
% EffiQA pioneers a for knowledge-intensive querying over structured knowledge. 
% Future research exploring multi-modal reasoning, advanced semantic matching, and dynamic model balancing could further enhance EffiQA's capabilities and efficiency. Ultimately, EffiQA represents a significant step toward enabling powerful yet computationally efficient knowledge-based AI through harmonious language model and knowledge graph synergies

% EffiQA framework enhances knowledge-based question answering by integrating LLMs with dynamic Knowledge Graphs for efficient and accurate responses. By leveraging smaller, specialized plug-in models for tasks like semantic pruning and deploying LLMs for global planning, EffiQA optimizes computational resources, improves operational efficiency, and maintains high accuracy across complex queries. This modular and scalable approach positions EffiQA as a solution that effectively eliminates LLM illusions and improves LLM output accuracy, providing new ideas for knowledge question answering.

% \bibliographystyle{abbrvnat} % 参考文献样式
% \bibliographystyle{unsrt} % 参考文献样式
\bibliography{custom} % 指定参考文献文件

\newpage
\appendix
\section{Alternative Pruning Strategy: Clustering-Based Approach}
\label{sec:appendix_clustering}

In addition to the classification-based semantic pruning employed in EffiQA, we investigated an alternative pruning strategy utilizing clustering as a tool model. This approach aimed to group semantically similar entities within the knowledge graph to streamline the exploration process. Specifically, we leveraged a BERT-based model to generate embeddings for each entity and applied K-Means clustering to partition these embeddings into distinct clusters. The hypothesis was that clustering could effectively reduce the search space by allowing the system to focus on the most relevant clusters based on the query context.

However, empirical evaluations on the WebQSP dataset revealed significant limitations of the clustering-based approach. The clustering model achieved an accuracy of approximately 20\%, which is substantially lower than the classification-based method integrated into EffiQA. This low accuracy can be attributed to several factors:

\begin{itemize}
    \item \textbf{Model Precision:} The BERT-based clustering model struggled to accurately group entities with nuanced semantic relationships, leading to high levels of intra-cluster heterogeneity.
    \item \textbf{Noise Sensitivity:} Clustering algorithms are inherently sensitive to noise and outliers, which are prevalent in large-scale knowledge graphs. This sensitivity resulted in the formation of clusters that contained a significant number of irrelevant entities.
    \item \textbf{Recall Rate:} The recall rate for the clustering-based approach was notably low, meaning that many relevant entities were inadvertently pruned during the clustering process. This shortfall undermined the effectiveness of the pruning strategy, as essential information required to answer complex queries was often excluded.
\end{itemize}

Furthermore, the presence of substantial noise and the inability to reliably distinguish between closely related entities significantly hindered the clustering model's performance. As a result, the clustering-based pruning strategy did not achieve the desired balance between efficiency and accuracy, making it an unsuitable alternative to the classification-based approach within the EffiQA framework.

\begin{figure}[ht]
  \centering
  \includegraphics[width=1\linewidth]{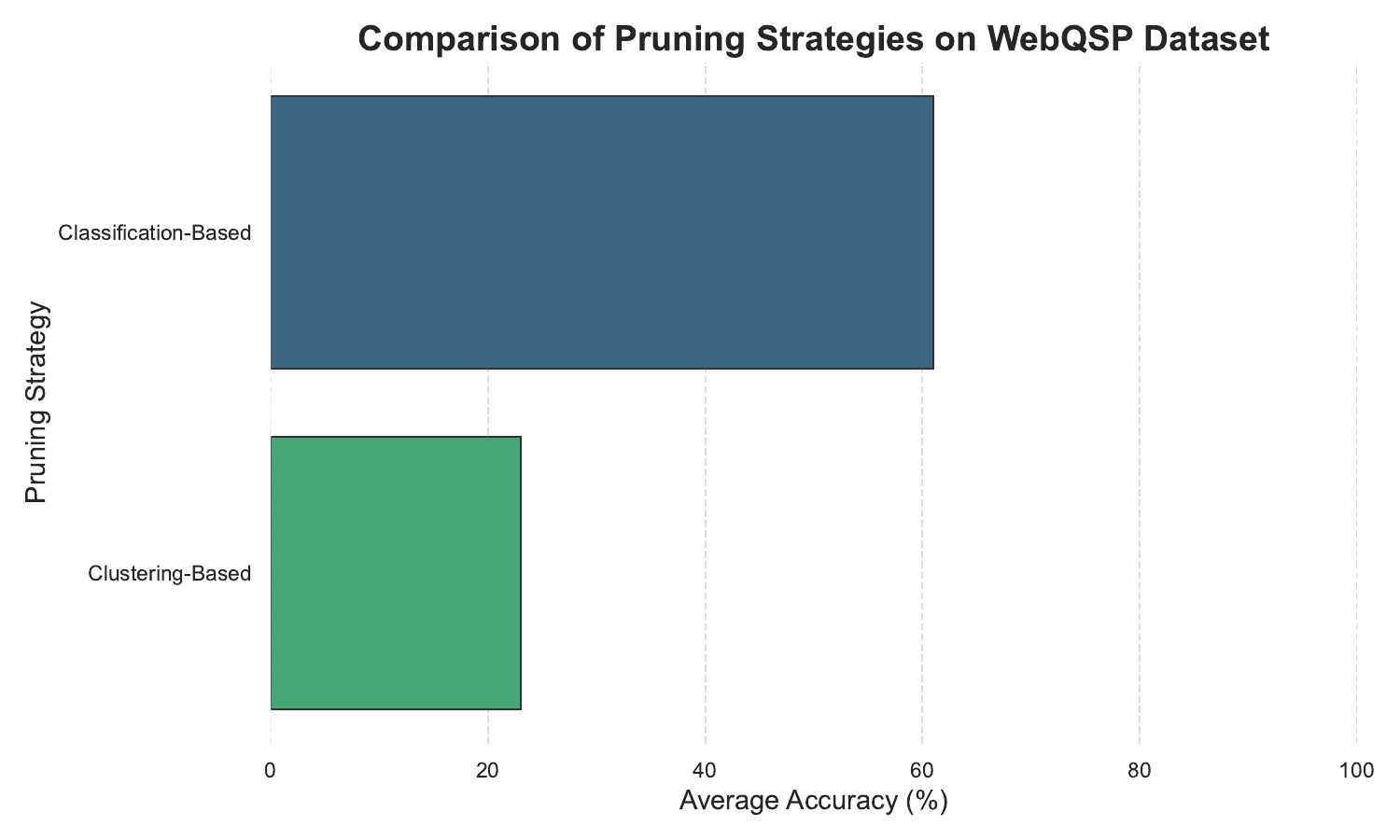} 
  \caption{Comparison of Classification and Clustering Accuracy on the WebQSP Dataset}
  \label{fig:clustering_vs_classification}
\end{figure}
To provide a visual comparison of the performance disparity between clustering and classification-based pruning, Figure \ref{fig:clustering_vs_classification} illustrates the accuracy differences observed on the WebQSP dataset.
\section{Case Study: Effective and Erroneous Query Processing}
\label{sec:appendix}

\subsection{Accurate Inference from a Structured Query}

\subsubsection{Query Example}
"Where did the 'Country Nation World Tour' concert artist go to college?"

\subsubsection{Analysis}

\textbf{Global Planning}

In this example, the Language Model (LLM) successfully identifies the primary entity: the artist associated with the 'Country Nation World Tour.' The model effectively decomposes the query into manageable components by first recognizing the main subject (the artist) and then focusing on a specific aspect of interest (their educational background).

\textbf{Search Execution}

With a clear global plan, the system initiates a targeted search within the knowledge graph. By concentrating on the identified artist, the LLM navigates the relevant nodes and edges efficiently to retrieve the associated college information.

\textbf{Answer Aggregation}

After gathering potential results, the system cross-verifies the information to ensure its consistency and relevance. This verification step confirms the accuracy of the retrieved college information, leading to a validated and coherent final answer.
\begin{figure*}[t]
  \centering
  \includegraphics[width=\textwidth]{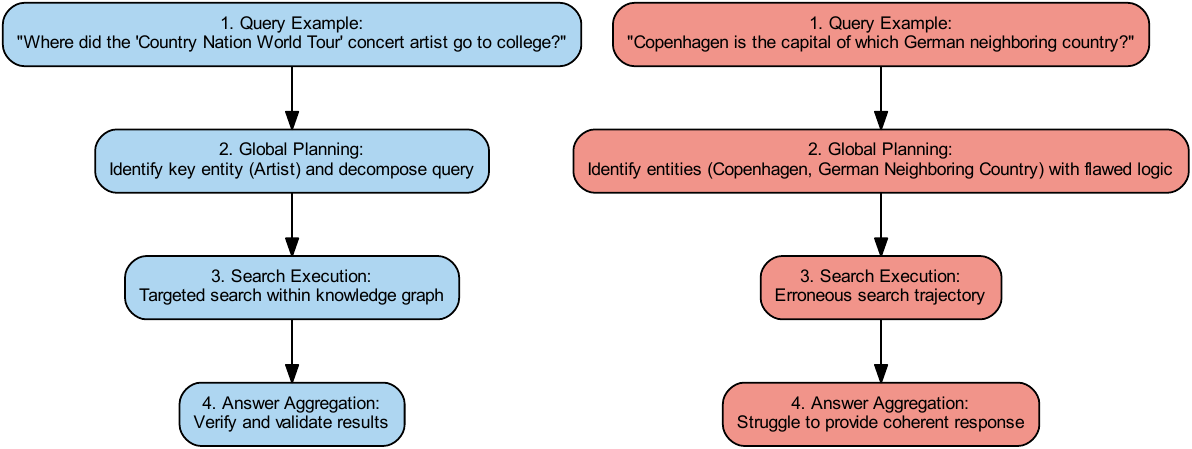} 
  \caption{Correct and incorrect example reasoning flow.}
  \label{fig:case_study}
\end{figure*}
\subsection{Error Analysis: Incorrect Query Framing}

\subsubsection{Query Example}
"Copenhagen is the capital of which German neighboring country?"

\subsubsection{Analysis}

\textbf{Global Planning}

In this case, the LLM identifies 'Copenhagen' and 'German neighboring country' as key entities. However, the logical connection within the query is flawed because Copenhagen is the capital of Denmark, not a German neighboring country. This misassociation leads to an incorrect understanding of the query's intent.

\textbf{Search Execution}

Following the flawed global plan, the LLM attempts to search for a German neighboring country associated with Copenhagen. This results in an erroneous search trajectory, as no such country exists, leading the system to retrieve invalid or irrelevant information.

\textbf{Answer Aggregation}

Due to the incorrect premise, the system struggles to aggregate a meaningful response. The lack of relevant results from the knowledge graph forces the system to produce an unconvincing or inaccurate answer, highlighting the limitation in handling logically flawed queries.

\subsubsection{Discussion}

These case studies highlight the importance of structured query processing in achieving accurate inference. In the first example, the system's ability to break down the query into actionable components led to successful information retrieval and accurate answer aggregation. Conversely, the second example exposes a significant limitation: while LLMs excel at identifying entities and executing searches, they can falter when confronted with queries that contain inherent logical errors or incorrect assumptions.

To mitigate such issues, enhancing the logical reasoning capabilities of LLMs is essential. Incorporating real-time error detection and query reformulation mechanisms could enable the system to identify and rectify flawed queries before executing searches. Future research may focus on integrating more advanced reasoning algorithms that can detect logical inconsistencies and guide the system in reformulating queries to improve overall robustness and accuracy.

\end{document}